%% file: acl_latex.tex
\documentclass[11pt]{article}

\usepackage[preprint]{acl}

\usepackage{times}
\usepackage{latexsym}
\usepackage{amsmath}

\usepackage[T1]{fontenc}
\usepackage[utf8]{inputenc}

\usepackage{microtype}

\usepackage{inconsolata}

\usepackage{graphicx}

\usepackage{bbm}
\usepackage{algorithm}
\usepackage{algpseudocode}
\usepackage{booktabs}

\algrenewcommand\algorithmicrequire{\textbf{Input:}}
\algrenewcommand\algorithmicensure{\textbf{Output:}}

\title{InfoMem: Training Long-Context Memory Agents with Answer-Conditioned Information Gain}

\author{
  \textbf{Tiancheng Han\textsuperscript{1,2}},
  \textbf{Yong Li\textsuperscript{1}},
  \textbf{Wuzhou Yu\textsuperscript{1}},
  \textbf{Qiaosheng Zhang\textsuperscript{2,3,\textdagger}},
  \textbf{Wenqi Shao\textsuperscript{2,3,\textdagger}}
\\
  \textsuperscript{1}Tongji University
  \textsuperscript{2}Shanghai Innovation Institute
  \textsuperscript{3}Shanghai AI Laboratory \\
  \texttt{zhangqiaosheng@pjlab.org.cn} \quad
  \texttt{weqish@link.cuhk.edu.hk}
}

\begin{document}
\maketitle
\begingroup
\renewcommand\thefootnote{}
\begin{NoHyper}
\footnotetext{\textsuperscript{\textdagger}Corresponding authors.}
\end{NoHyper}
\endgroup
\begin{abstract}
Long-context tasks require LLMs to identify and preserve answer-relevant information 
from large contexts. 
Chunk-wise memory agents address this issue by sequentially reading document chunks, 
updating a compact memory, and generating the final answer from the accumulated memory. 
However, existing RL-based chunk-wise agents either rely on sparse final-answer rewards 
or use lexical intermediate rewards for memory and retrieval actions. These signals 
supervise task success or local overlap, but do not directly evaluate whether the final 
memory supports the ground-truth answer.
We propose InfoMem, a reward mechanism for training chunk-wise memory agents 
that evaluates final-memory utility using answer-conditioned information. 
InfoMem measures how much the final memory increases the model's per-token 
log-likelihood of the ground-truth answer.
To stabilize RL optimization, InfoMem applies this signal only to successful 
trajectories and normalizes it before reward composition.
Under the same GRPO framework and training budget, InfoMem improves 
long-context memory-agent performance over comparable 
memory-agent RL baselines. 
Analyses show that effective final-memory rewards should operate 
on successful trajectories, be normalized before reward composition, 
and be conditioned on the answer rather than the query. 
Our code is available at \url{https://github.com/GenSouKa1/InfoMem}.
\end{abstract}

\section{Introduction}

Long-context understanding has become a central capability for large language models (LLMs), 
with applications ranging from long-document question answering to corpus-level 
evidence aggregation~\citep{openai2025mrcr,wulongmemeval,hsiehruler,lu2026corpusqa10milliontoken}. 
Prior work has improved long-context processing through extended context 
windows~\citep{shen2025qwenlongl15posttrainingrecipelongcontext}, 
attention or positional modifications~\citep{munkhdalai2024leavecontextbehindefficient,presstrain}, 
retrieval augmentation~\citep{zhao2024longrag}, 
and memory-based or agentic 
pipelines~\citep{packer2024memgptllmsoperatingsystems,zhou2025mem1learningsynergizememory}. 
However, effectively using long inputs remains challenging when relevant evidence is sparse, 
distributed across distant segments, or must be preserved throughout a long reading process.

Among these approaches, chunk-wise memory agents provide a simple and effective 
paradigm for long-context reasoning. 
Instead of processing the entire document at once, the model sequentially reads 
shorter chunks, updates a compact memory state, and generates 
the final answer from the accumulated memory. 
This paradigm appears in training-free reading agents~\citep{lee2024human}, 
recurrent or memory-augmented model architectures~\citep{dai-etal-2019-transformer}, 
and post-training methods~\citep{yu2025memagentreshapinglongcontextllm,shi2026lookreasonforwardrevisitable}. 
Crucially, their explicit memory state makes memory 
quality an observable optimization target.

Despite their effectiveness, existing chunk-wise long-context systems still lack a 
scalable way to supervise memory formation.
Training-free methods often rely on human-designed memory-update prompts, 
summarization heuristics, or fixed traversal workflows~\citep{lee2024human}. 
Architecture-level methods may improve long-context capability more fundamentally
~\citep{dai-etal-2019-transformer}, 
but typically require costly pretraining.
Reinforcement learning (RL)-based memory agents can also improve long-context behavior 
through task feedback, 
but existing methods mainly rely on sparse answer rewards~\citep{yu2025memagentreshapinglongcontextllm} 
or lexical intermediate rewards for memory and retrieval actions~\citep{shi2026lookreasonforwardrevisitable}. 
These rewards supervise task success or local word overlap, without directly evaluating 
whether the final memory semantically supports the ground-truth answer.

This limitation is especially pronounced within successful trajectories: sparse outcome 
rewards cannot distinguish whether the final memory contains focused answer-supporting 
evidence or redundant distracting information, while lexical rewards may not capture 
semantic support for the final answer.
This motivates a memory-specific reward signal for chunk-wise long-context reinforcement learning.

We propose InfoMem, a reward-shaping method for chunk-wise long-context memory agents 
based on answer-conditioned information gain. 
The core intuition is that a useful final memory should increase the model's support 
for the ground-truth answer. 
Instead of estimating distribution-level mutual information, InfoMem uses a pointwise 
information-gain surrogate by comparing the model's per-token log-likelihood of 
the ground-truth answer with and without the final memory. 
InfoMem further improves training stability by applying this signal only to 
successful trajectories and normalizing it before reward composition.

Experiments show that InfoMem consistently improves long-context memory-agent 
performance over outcome-only GRPO and a comparable memory-agent RL baseline 
ReMemR1~\citep{shi2026lookreasonforwardrevisitable}. 
Further analyses show that effective final-memory rewards should operate on 
successful trajectories, be normalized before reward composition, and be 
conditioned on the ground-truth answer rather than on the query alone. 
These findings suggest that answer-conditioned information gain provides a 
principled framework for final-memory supervision in chunk-wise long-context RL.

Our contributions are threefold: (1) We formulate final-memory utility through an 
information-theoretic perspective, where useful memories should reduce the model's uncertainty 
about the ground-truth answer. (2) We introduce InfoMem, an answer-conditioned information-gain 
reward for direct final-memory shaping over successful trajectories. (3) We show that InfoMem 
consistently improves chunk-wise long-context memory agents over comparable memory-agent RL baselines, 
and further identify three key properties of effective final-memory rewards: 
success-side supervision, pre-composition normalization, and answer conditioning.

\section{Related Work}

\subsection{Long-context LLMs and Chunk-wise Memory Agents}

Long-context LLM research has improved long-input processing 
through context extension~\citep{shen2025qwenlongl15posttrainingrecipelongcontext}, 
attention or positional 
modifications~\citep{munkhdalai2024leavecontextbehindefficient,presstrain}, 
and retrieval augmentation~\citep{zhao2024longrag}. 

Complementary to these approaches, chunk-wise memory agents 
maintain an explicit memory state during sequential long-document processing. 
This paradigm includes training-free reading workflows~\citep{lee2024human}, 
segment-level recurrent 
architectures~\citep{dai-etal-2019-transformer,NEURIPS2022_d6e0bbb9,ding2021ernie}, 
and post-training memory 
agents~\citep{yu2025memagentreshapinglongcontextllm,shi2026lookreasonforwardrevisitable}. 
Existing methods rely on manually designed workflows, costly recurrent 
architectures, or RL objectives based on sparse final-answer rewards and 
intermediate memory heuristics. 
As a result, direct supervision of answer-conditioned final-memory utility 
remains relatively underexplored.

\subsection{RL for Long-context QA}

DeepSeek-R1~\citep{guo2025deepseek} demonstrates that reinforcement learning 
can substantially enhance LLM capabilities in specialized domains. 
Recent work further shows that RL can improve long-context question answering and reasoning. 
Early methods mainly optimize verifiable end-task outcomes, such as final-answer 
correctness or verifier-based response 
quality~\citep{shen2025qwenlongl15posttrainingrecipelongcontext,yu2025memagentreshapinglongcontextllm}. 
More recent studies have introduced denser supervision signals for grounding, 
evidence extraction, and 
contextual reasoning~\citep{chenlongrlvr,guan2026evidenceaugmentedpolicyoptimizationreward,
ping2026longrunleashinglongcontextreasoning,shi2026lookreasonforwardrevisitable}. 
Together, these results suggest that reward-based optimization is a 
promising direction for improving long-context reasoning.

Despite these advances, existing rewards mainly supervise grounding quality, 
evidence selection, reading utility, or intermediate memory-update behavior. 
ReMemR1~\citep{shi2026lookreasonforwardrevisitable} introduces 
information-style rewards for memory and callback actions, but these 
signals are based on word-level recall rather than answer-conditioned final-memory utility. 
Direct supervision of the final memory representation itself remains relatively underexplored. 
In particular, existing methods rarely evaluate whether the resulting 
final memory directly supports the ground-truth answer, which is the focus of our work.

\section{Problem Setup and Motivation}

\subsection{Chunk-wise Long-context Memory Agent}
\label{sec:chunk-wise-agent}

We focus on chunk-wise memory agents as a practical paradigm for reinforcement learning in 
long-context settings. Formally, let $x$ denote the query, $D$ the long document, 
and $y^*$ the ground-truth answer. Given a pre-defined chunk size $C$, 
the document is divided into $K$ chunks:
\begin{equation}
  D = \{c_1, c_2, \ldots, c_K\},
\end{equation}
Conditioned on the query, the model sequentially reads the chunks and maintains a memory state:
\begin{equation}
  M_t = \pi_\theta(M_{t-1}, c_t, x), \quad t = 1,\ldots,K.
\end{equation}
After processing all chunks, the model obtains the final memory $M_K$ and generates the final answer based on the query and this memory:
\begin{equation}
  \hat{y} = \pi_\theta(x, M_K).
\end{equation}

\subsection{Why Outcome Reward Is Insufficient for Memory Learning}
\label{sec:outcome-insufficient}

The outcome reward directly supervises final answer correctness, 
but provides only sparse and indirect supervision for memory utility. 
In chunk-wise memory agents, the final prediction is generated based on the final memory $M_K$, 
which is expected to retain the information necessary for supporting the correct answer. 
However, the binary outcome reward evaluates only whether the generated answer matches $y^*$, 
without directly distinguishing the quality of different final memories.

This limitation is particularly evident among successful trajectories. 
Multiple rollouts may generate the same correct answer and therefore receive identical 
outcome rewards, while their final memories can differ substantially in utility. 
Some memories may retain only the critical supporting evidence, whereas others may preserve 
the same evidence together with substantial redundant or distracting information while still 
yielding the correct prediction. Consequently, outcome reward alone cannot differentiate 
memory utility within successful trajectories, motivating the need for a direct reward 
signal for final-memory utility in chunk-wise long-context reinforcement learning.

\subsection{From Mutual Information to Model-induced Pointwise Surrogate}
\label{sec:mi-surrogate}

Long-context question answering can be viewed as extracting answer-relevant information 
from a large context. Under the chunk-wise memory-agent formulation, the final memory 
should therefore reduce the uncertainty of the answer conditioned on the query. 
Ideally, this utility can be characterized by the conditional mutual information 
$I(M;Y \mid X)$, which measures how much additional information the memory $M$ provides 
about the answer $Y$ given the query $X$.

However, distribution-level mutual information depends on the full joint distribution 
over queries, memories, and answers, and is difficult to estimate reliably in the 
high-dimensional semantic space of LLMs~\citep{qian2026demystifying}. 
This motivates the InfoMem reward introduced in Section~\ref{sec:method}, where we 
instantiate the mutual-information objective with a single-sample pointwise surrogate. 
This surrogate measures whether the final memory increases the model's support for the 
ground-truth answer on the current instance.

\section{Method: InfoMem}
\label{sec:method}

\begin{figure*}[t]
  \centering
  \includegraphics[width=\textwidth]{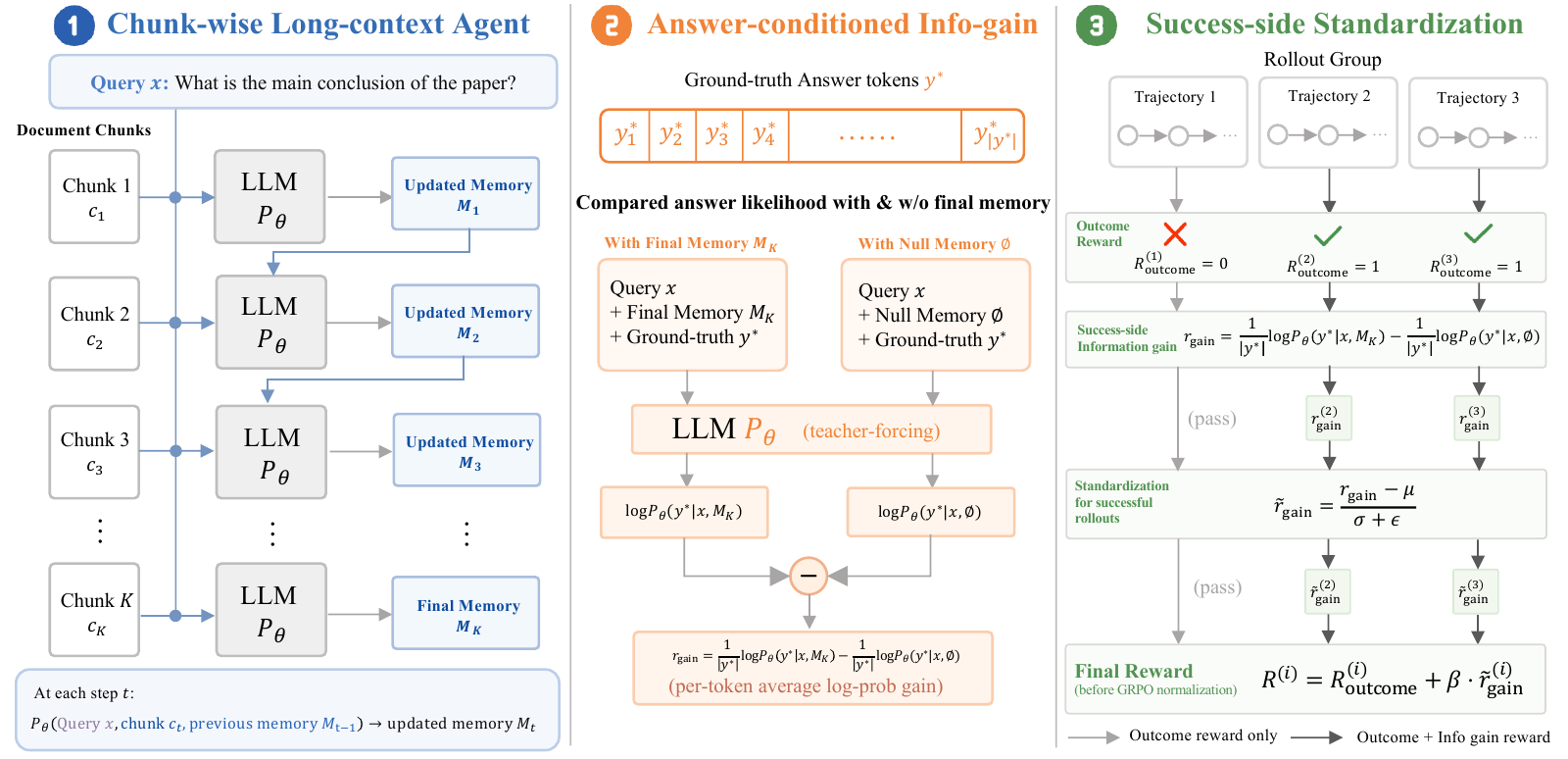}
  \caption{Overview of InfoMem for chunk-wise long-context RL. 
  InfoMem measures final-memory utility using answer-conditioned information gain by 
  comparing the teacher-forced per-token average log-likelihood of the ground-truth 
  answer $y^*$ with and without the final memory. During GRPO training, 
  information-gain supervision is applied only to successful trajectories, 
  normalized across successful rollouts, and combined with the sparse outcome 
  reward for policy optimization.}
  \label{fig:infomem}
\end{figure*}

\subsection{Reward Definition}
\label{sec:reward-definition}

We train the chunk-wise memory agent with Group Relative Policy Optimization (GRPO)~\citep{shao2024deepseekmathpushinglimitsmathematical}. 
For each prompt, GRPO samples a  group of $n$ rollouts,
\begin{equation}
  \mathcal{G} = \{1,2,\ldots,n\}.
\end{equation}
Each rollout $i \in \mathcal{G}$ produces a final memory $M_i$ and a final answer 
$\hat{y}_i$. The base outcome reward is defined as
\begin{equation}
  R_{\mathrm{outcome}, i}
  =
  \mathbbm{1}[\hat{y}_i = y^*],
\end{equation}
where answer correctness is evaluated by normalized string matching in our main 
experiments.

Given a query $x$, a final memory $M$, and the 
ground-truth answer $y^*$, we define the answer-conditioned information-gain reward 
$r_{\mathrm{gain}}$ as
\begin{equation}
  \resizebox{0.98\columnwidth}{!}{$\displaystyle
  r_{\mathrm{gain}}(x,M,y^*)
  =
  \frac{1}{|y^*|}\log P_\theta(y^* \mid x,M)
  -
  \frac{1}{|y^*|}\log P_\theta(y^* \mid x,\emptyset),
  $}
\label{eq:rgain}
\end{equation}
where $\emptyset$ denotes null memory. The sequence likelihood of LLM on ground-truth tokens 
$P_\theta$ is computed under teacher forcing:
\begin{equation}
\log P_\theta(y^* \mid x,M)
=
\sum_{j=1}^{|y^*|}
\log p_\theta
(
y^*_j
\mid
y^*_{<j},x,M
),
\label{eq:logP}
\end{equation}
which corresponds to the summed log-probability of the ground-truth tokens.
The first term in Eq.~\eqref{eq:rgain} measures the likelihood assigned to the ground-truth 
answer when conditioned on the final memory, while the second term measures the 
corresponding likelihood without memory conditioning. Consequently, higher values of 
$r_{\mathrm{gain}}$ indicate that the final memory increases the likelihood assigned to 
the correct answer.

% Equivalently, Eq.~\eqref{eq:rgain} can be written as a difference in 
% per-token average cross-entropy:

% \begin{equation}
%   \resizebox{0.92\columnwidth}{!}{$\displaystyle
%   r_{\mathrm{gain}}(x,M,y^*)
%   =
%   -\mathrm{CE}_\theta(y^* \mid x,M)
%   +
%   \mathrm{CE}_\theta(y^* \mid x,\emptyset),
%   $}
% \label{eq:rgain_ce}
% \end{equation}
% where $\mathrm{CE}_\theta$ denotes the per-token cross-entropy loss under teacher forcing.

\subsection{Using Successful Trajectories as Positive Memory Signals}
\label{sec:success-side-supervision}

Given the rollout group $\mathcal{G}$ and the outcome rewards defined above, 
we define the set of successful rollouts as
\begin{equation}
  \mathcal{S}
  =
  \{i \in \mathcal{G}: R_{\mathrm{outcome},i}=1\}.
  \label{eq:success_rollouts}
\end{equation}

InfoMem applies the information-gain reward only to rollouts in $\mathcal{S}$. 
Specifically, successful trajectories are further differentiated by 
$r_{\mathrm{gain}}(x,M_i,y^*)$, whereas failed trajectories are supervised 
with $R_{\mathrm{outcome}}$ only.

This design restricts information-gain optimization to trajectories whose final 
answers are already validated by the outcome reward. Within this subset, differences 
in $r_{\mathrm{gain}}$ more directly reflect the extent to which the final memory 
supports the ground-truth answer $y^*$, thereby providing a more stable signal for 
memory shaping.

By contrast, failed trajectories may entangle memory quality with incorrect evidence 
selection or answer generation, making $r_{\text{gain}}$ less reliable as a memory-utility 
signal. We empirically analyze this issue through wrong-only and 
both-side supervision variants in Section~\ref{sec:analysis}.

\subsection{Memory Reward Normalization}
\label{sec:success-side-normalization}

The scale of raw information-gain rewards can vary substantially across prompts due to 
differences in answer uncertainty. For relatively easy questions, 
the model may already assign high likelihood to the ground-truth answer 
without memory conditioning, resulting in limited likelihood improvement from the final memory. 
For more difficult questions, an informative final memory can produce a 
substantially larger increase in ground-truth likelihood. 
Consequently, directly combining raw $r_{\mathrm{gain}}$ with the binary outcome reward 
may introduce severe reward-scale imbalance, causing the information-gain term to dominate 
the sparse outcome signal for some prompts while remaining negligible for others.

Following the reward-decoupled normalization setting in 
GDPO~\citep{liu2026gdpogrouprewarddecouplednormalization}, 
InfoMem controls this scale before reward composition by normalizing information-gain 
values within the successful trajectories of the same rollout group. Let
$r_i = r_{\mathrm{gain}}(x,M_i,y^*)$ for $i \in \mathcal{S}$. We compute
\begin{equation}
  \tilde{r}_i
  =
  \frac{r_i - \mu_{\mathcal{S}}}{\sigma_{\mathcal{S}}+\epsilon},
  \quad i \in \mathcal{S},
  \label{eq:success_normalization}
\end{equation}
where $\mu_{\mathcal{S}}$ and $\sigma_{\mathcal{S}}$ denote the mean and 
standard deviation of $\{r_i : i \in \mathcal{S}\}$, and $\epsilon$ is a small constant 
for numerical stability. When $|\mathcal{S}|=0$, no information-gain reward is applied. 
When $|\mathcal{S}|=1$, the original value is retained as a fallback strategy.

The final reward for rollout $i$ is defined as
\begin{equation}
  R_i
  =
  \begin{cases}
  R_{\mathrm{outcome},i} + \beta \tilde{r}_i, & i \in \mathcal{S},\\
  R_{\mathrm{outcome},i}, & i \notin \mathcal{S},
  \end{cases}
  \label{eq:combined_reward}
\end{equation}
where $\beta$ controls the strength of the normalized information-gain term.

This normalization differs from the group-relative normalization in GRPO. 
InfoMem normalizes $r_{\mathrm{gain}}$ before reward composition, controlling the 
relative scale between the information-gain and outcome rewards. 

Although InfoMem is computed from the final memory and final answer, the resulting 
trajectory-level advantage is propagated to all generated tokens, including both 
memory-update and final-answer tokens. 
The complete reward-construction procedure is summarized in 
Appendix~\ref{app:grpo-reward-implementation}.

\section{Experiments}

\subsection{Synthetic Context Discrimination with Hallucinated Evidence}
\label{sec:synthetic-hallucination}

We first evaluate whether answer-conditioned information gain can distinguish 
genuinely answer-supporting evidence from surface-similar but factually misleading contexts. 
We construct a synthetic diagnostic using the SQuAD dataset~\citep{rajpurkar-etal-2016-squad}. 
Each example contains a question, a supporting context, and a ground-truth answer. 

\begin{figure}[t]
  \centering
  \includegraphics[width=\columnwidth]{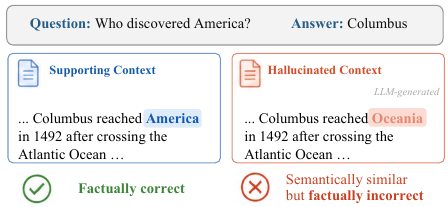}
  \caption{Example of synthetic hallucinated evidence.}
  \label{fig:hallucination}
\end{figure}

For each retained QA pair, we generate two hallucinated contexts using 
Gemini 3 Flash preview~\citep{google2026gemini3flashpreview}. 
The hallucinated contexts preserve the overall structure and semantic style of the original 
passage while replacing answer-critical facts so that they no longer support the 
ground-truth answer, as illustrated in Figure~\ref{fig:hallucination}. 

We compare three categories of evidence scores: 
(1) \textbf{Information gain.} We compute the $r_{\mathrm{gain}}$ score using 
the same Qwen2.5-1.5B-Instruct model~\citep{qwen2.5technicalreport} as in training, 
comparing the per-token average log-likelihood of the ground-truth answer with and 
without the candidate context. 
(2) \textbf{Embedding similarity.} We compute cosine similarity between the embedding of 
candidate-context and question-answer template, using 
mainstream open-source embedding models~\citep{zhang2025qwen3embeddingadvancingtext,chen-etal-2024-m3,wang2024multilinguale5}. 
(3) \textbf{Attention-based scores.} Using the same Qwen2.5-1.5B-Instruct model, 
we evaluate two attention metrics. 
\textsc{Attn-Mass} averages the total attention mass assigned from answer tokens 
to context tokens, while \textsc{Attn-Top1} averages the maximum attention score 
assigned to any context token for each answer token.

We evaluate each score from two complementary perspectives. 
First, we compute the mean reciprocal rank (MRR) of the true supporting context among the three 
candidates. 
Second, we report a Z-score signal-to-noise ratio (SNR) to assess the stability of this 
discrimination signal. 
A higher SNR indicates that the score not only separates true contexts from hallucinated ones, 
but also does so with lower relative variation across samples, 
making it more suitable as a training reward. 

\begin{table}[t]
\centering
{\small
\setlength{\tabcolsep}{4pt}
\renewcommand{\arraystretch}{1.15}
\begin{tabular}{@{}lcc@{}}
\hline
\textbf{Score} & \textbf{MRR} & \textbf{Z-score SNR} \\
\hline
Embedding (BGE-M3) & 0.719 & 0.316 \\
Embedding (E5) & 0.714 & 0.320 \\
Embedding (Qwen3-0.6B) & 0.727 & 0.371 \\
\textsc{Attn-Mass} & 0.588 & -0.081 \\
\textsc{Attn-Top1} & 0.792 & 0.577 \\
$r_{\text{gain}}$ (ours) & \textbf{0.977} & \textbf{2.960} \\
\hline
\end{tabular}
}
\caption{Synthetic context discrimination results. The $r_{\text{gain}}$ score corresponds 
to the answer-conditioned information-gain score in Eq.~\eqref{eq:rgain}.}
\label{tab:synthetic-context-discrimination}
\end{table}

\begin{table*}[t]
\centering
{\small
\setlength{\tabcolsep}{6pt}
\renewcommand{\arraystretch}{1.15}
\begin{tabular}{lcccc}
\hline
\textbf{Model} & \textbf{CorpusQA} & \textbf{LongMemEval} & \textbf{MRCR-8needle} & \textbf{RULER synthetic QA} \\
\hline
Qwen2.5-1.5B-Instruct & 14.590 & 5.600 & 0.260 & 13.308 \\
with Outcome-only GRPO  & 16.413 & 10.000 & 0.063 & 34.735 \\
with InfoMem & \textbf{19.453} & \textbf{12.800} & \textbf{0.279} & \textbf{36.848} \\
\specialrule{0.15pt}{0pt}{0pt}
with ReMemR1~\citep{shi2026lookreasonforwardrevisitable} & 1.520 & 6.200 & 0.092 & 28.919 \\
\hline
\end{tabular}
}
\caption{Main evaluation results on long-context benchmarks. All scores are reported as 
percentages. The best score for each benchmark is shown in bold.}
\label{tab:main-evaluation}
\end{table*}

Table~\ref{tab:synthetic-context-discrimination} shows that 
$r_{\text{gain}}$ achieves the highest MRR and Z-score SNR among all compared scores. 
The high MRR indicates strong discrimination between true and hallucinated contexts, 
while the substantially larger SNR suggests that the resulting signal is also more stable 
across samples. Embedding-based similarities achieve moderate ranking performance but 
exhibit much weaker SNR. Attention-based scores, especially \textsc{Attn-Top1}, improve 
ranking quality but remain substantially below $r_{\text{gain}}$ in signal stability. 
These results suggest that answer-conditioned likelihood gain provides both stronger 
evidence discrimination and a cleaner reward signal for training.
Details of the discrimination experiment are provided in 
Appendix~\ref{app:synthetic-hallucination}.

\subsection{Training Setup}
\label{sec:training-setup}

We train on the RULER-HotpotQA dataset~\citep{yu2025memagentreshapinglongcontextllm}, 
which applies the RULER long-context construction paradigm~\citep{hsiehruler} to 
HotpotQA~\citep{yang-etal-2018-hotpotqa} by mixing answer-relevant documents with 
distractors. Each example contains 200 documents. The original training split has 
32,768 examples and the validation split has 128 examples; for controlled experimentation, 
we downsample the training split to 512 examples and keep the validation split unchanged.

All experiments use Qwen2.5-1.5B-Instruct as the base model and GRPO as the algorithm. 
For each prompt, we sample $n=8$ rollouts and train for 120 steps with 
learning rate $1\times10^{-6}$. The chunk size is 5000 tokens, the maximum memory length 
is 1024 tokens, and the InfoMem coefficient is $\beta=0.2$. 
For controlled comparison, all methods use the same data, model, rollout number, decoding 
configuration, and training budget. Our primary baseline is Outcome-only GRPO~\citep{yu2025memagentreshapinglongcontextllm}, 
which removes the information-gain term while keeping the same training pipeline. We also report 
the initial model before RL optimization where appropriate. 
Since ReMemR1~\citep{shi2026lookreasonforwardrevisitable} augments the chunk-wise 
memory-agent framework with callback retrieval, we align all shared training parameters 
with our setting for fair reproduction, while keeping ReMemR1-specific parameters 
consistent with the original paper.
Details are provided in Appendix~\ref{app:implementation-details}.

\subsection{Evaluation}
\label{sec:evaluation}

\subsubsection{Benchmarks}
\label{sec:benchmarks}

Under the same chunk-wise memory framework, we evaluate the trained models on a set 
of out-of-distribution long-context benchmarks covering complementary forms of evidence use:
(1) \textbf{MRCR-8needle}~\citep{openai2025mrcr} evaluates multi-needle retrieval; 
(2) \textbf{RULER synthetic QA}~\citep{yu2025memagentreshapinglongcontextllm} evaluates 
sparse retrieval-style question answering, using a different corpus source from the 
training data; 
(3) \textbf{CorpusQA}~\citep{lu2026corpusqa10milliontoken} evaluates corpus-level evidence 
aggregation across document collections; and 
(4) \textbf{LongMemEval}~\citep{wulongmemeval} evaluates long-horizon dialogue memory and 
conversational information tracking.

Across benchmarks, we use the metric specified by each benchmark and apply the same 
evaluation protocol to all compared methods, except that ReMemR1 is evaluated under 
its original chunk-wise framework with callback retrieval for train-test consistency. CorpusQA and LongMemEval are evaluated 
with LLM-as-judge using Kimi-K2.6~\citep{moonshot2026k2_6}\footnote{Manual verification 
shows high similarity to human judgments: 99.5\% on CorpusQA and 96.8\% on LongMemEval.}, MRCR-8needle with sequence match, and RULER synthetic QA with F1 score. 
For evaluation efficiency, we downsample the original 800 MRCR-8needle examples to 
100 examples, use the 128K-token subset of CorpusQA, and use the 115K-token 
LongMemEval-S subset.
Details for evaluation are provided in Appendix~\ref{app:evaluation-protocols}.

\subsubsection{Main Results}
\label{sec:main-results}

Table~\ref{tab:main-evaluation} shows that InfoMem achieves the best overall performance across 
all four long-context benchmarks, outperforming both Outcome-only GRPO and ReMemR1. 
Compared with the initial model, 
InfoMem consistently improves both retrieval-oriented tasks and more memory-intensive settings. 
These results suggest that answer-conditioned information gain provides an effective supervision 
signal for final-memory formation beyond sparse final-answer correctness.

In contrast, Outcome-only GRPO produces less consistent improvements. Although it improves 
over the initial model on CorpusQA, LongMemEval, and RULER synthetic QA, it substantially 
degrades performance on MRCR-8needle, falling below the initial model. 
This result suggests that optimizing only sparse outcome rewards may lead to degraded 
long-context retrieval behavior. InfoMem avoids this degradation and achieves the best overall 
performance, supporting the need for a memory-specific reward signal in chunk-wise long-context RL.

ReMemR1 also fails to match InfoMem under our sample-aligned evaluation. 
One possible explanation is that its intermediate reward design is primarily 
based on word-level recall against the ground-truth answer, 
which may encourage lexical overlap rather than preserving semantically 
supporting evidence.

\section{Analysis and Ablation}
\label{sec:analysis}

We next analyze the key design choices underlying InfoMem. 
Specifically, we study which trajectories should receive information-gain supervision, 
why information-gain rewards must be normalized before reward composition, 
and why the reward should be conditioned on the ground-truth answer rather than on the query.

\subsection{Which Side Should Receive Information-gain Supervision?}
\label{sec:supervision-side}

We analyze which trajectories should receive information-gain supervision. 
Specifically, we compare three supervision-side variants: 
\textit{Success} (the default InfoMem setting);
\textit{Wrong}, which applies $r_{\mathrm{gain}}$ only to failed trajectories; 
and \textit{Both}, which applies it to both. 
All variants use the same training setup, differing only in the supervision.

\begin{figure}[t]
  \centering
  \includegraphics[width=\columnwidth]{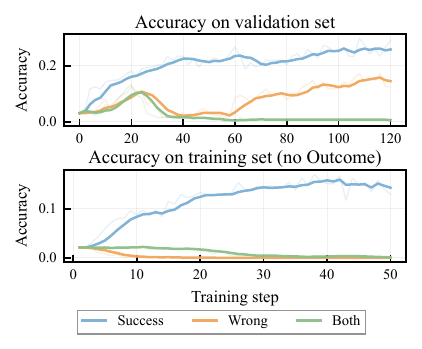}
  \caption{
Effect of information-gain supervision side. Top: with outcome reward, 
validation accuracy measures generalization. Bottom: without explicit outcome reward, 
training accuracy diagnoses whether each supervision side provides a usable learning signal.
Thick lines show sliding-window smoothing with window size 5, and light lines show raw values.
}
  \label{fig:supervision-side-ablation}
\end{figure}

\begin{table*}[t]
\centering
{\small
\setlength{\tabcolsep}{6pt}
\renewcommand{\arraystretch}{1.15}
\begin{tabular}{lcccc}
\hline
\textbf{Model} & \textbf{CorpusQA} & \textbf{LongMemEval} & \textbf{MRCR-8needle} & \textbf{RULER synthetic QA} \\
\hline
InfoMem & \textbf{19.453} & \textbf{12.800} & \textbf{0.279} & \textbf{36.848} \\
InfoMem w/o $r_{\mathrm{gain}}$ normalization & 16.109 & 11.600 & 0.029 & 35.371 \\
InfoMem with QueryPMI & 18.237 & 8.000 & 0.117 & 26.163 \\
\hline
\end{tabular}
}
\caption{Ablation results on long-context benchmarks. All scores are reported as 
percentages. The best score for each benchmark is shown in bold.}
\label{tab:ablation-evaluation}
\end{table*}

\subsubsection{Main Setting with Outcome Reward}

We retain the outcome reward and vary only the supervision side of the information-gain term. 
As shown in the top panel of Figure~\ref{fig:supervision-side-ablation}, 
success-only supervision produces the most stable validation.
This behavior supports the role of successful trajectories as positive memory examples: 
their final memories are already associated with correct final answers, and 
$r_{\mathrm{gain}}$ can further distinguish which successful memories provide stronger 
support for the ground-truth answer.

By contrast, wrong-side and both-side supervision exhibit unstable training dynamics. 
Both variants initially improve but collapse during mid-stage training, 
with the both-side variant remaining near zero afterward. 
This suggests that failed trajectories may provide weak useful signals 
early in training, but their associated memories become increasingly noisy 
as optimization proceeds. 
Mixing successful and failed memories within the same reward objective 
further weakens the positive memory-shaping effect of success-only supervision.
\subsubsection{No-outcome Diagnostic}

We further conduct a diagnostic experiment without explicit outcome reward. 
Rather than serving as the main training objective, this experiment tests whether 
$r_{\mathrm{gain}}$ alone can provide a usable learning signal under different supervision sides. 
Specifically, we remove $R_{\mathrm{outcome}}$ and optimize using only the 
information-gain reward under success-only, wrong-only, and both-side supervision.

The bottom panel of Figure~\ref{fig:supervision-side-ablation} shows that only success-only 
supervision produces sustained improvement in training accuracy, whereas wrong-only and 
both-side supervision fail to produce stable learning behavior. 
This result suggests that success-side information gain can provide meaningful memory shaping 
even without explicit outcome reward, while wrong-side supervision does not provide a reliable 
optimization signal. 
Overall, these findings support using information-gain reward as positive memory shaping over 
successful trajectories rather than as uniform supervision across all trajectories.

\subsection{Why Is Normalization Necessary Before Reward Composition?}
\label{sec:normalization-ablation}

We evaluate the role of pre-composition normalization by removing the normalization of 
$r_{\mathrm{gain}}$ before reward composition while keeping all other settings unchanged. 
As shown in Table~\ref{tab:ablation-evaluation}, removing this normalization consistently 
degrades performance across all evaluated benchmarks. 
The degradation is especially pronounced on retrieval-oriented tasks, suggesting that raw 
likelihood-gain rewards are not directly comparable across rollout groups.

These results support the need for normalization before reward composition. 
Since raw $r_{\mathrm{gain}}$ magnitudes can vary substantially across prompts, directly 
combining them with sparse outcome rewards may introduce unstable reward scaling. 
Normalizing $r_{\mathrm{gain}}$ within successful trajectories converts it into a 
relative memory-quality signal before reward composition, producing a more stable 
reward-shaping signal during training.

\subsection{Why Must the Reward Be Answer-conditioned Rather Than Query-conditioned?}
\label{sec:querypmi-ablation}

\begin{figure}[t]
  \centering
  \includegraphics[width=\columnwidth]{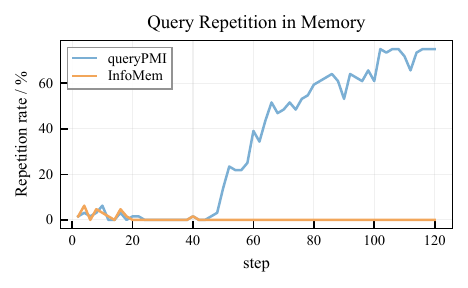}
  \caption{Fraction of rollouts whose final memory repeats the query during training.}
  \label{fig:querypmi}
\end{figure}

We next examine whether query relevance alone is sufficient for final-memory reward design. 
To this end, we construct a query-conditioned variant, QueryPMI, which replaces the 
ground-truth answer in $r_{\mathrm{gain}}$ with the query itself. 
Specifically, given a final memory $M$ and the query $x$, we use the template
\textit{``Based on the previous memory \{M\}, 
we can answer the query \{x\}.''}
and compute the corresponding per-token log-likelihood gain on query.
This formulation is intuitively appealing because useful memories are often related to the query. 
However, query relevance is not equivalent to answer support: a memory can increase 
query likelihood simply by copying or paraphrasing the query without preserving answer-supporting 
evidence.

Table~\ref{tab:ablation-evaluation} shows that QueryPMI performs consistently worse than InfoMem 
across all evaluated benchmarks, indicating that query-conditioned likelihood gain is a 
substantially weaker final-memory reward than answer-conditioned information gain.
Figure~\ref{fig:querypmi} further explains this degradation. 
During training, QueryPMI rapidly increases the proportion of rollouts whose final memories repeat 
the query, whereas InfoMem keeps this proportion close to zero. 
This behavior suggests that the model exploits the query-conditioned objective by making 
the query easier to predict from memory rather than learning to preserve answer-supporting evidence. 
These results highlight the importance of conditioning the reward on the ground-truth 
answer rather than on the query alone.
Additional training and validation curves are provided in 
Appendix~\ref{app:training-curves-diagnostics}.

\section{Conclusion}

We studied final-memory reward design for chunk-wise long-context memory agents. 
While outcome-only GRPO supervises memory formation only indirectly through final-answer correctness, 
InfoMem introduces answer-conditioned information gain as a direct reward signal for final-memory utility. 
Experiments show that InfoMem consistently improves long-context memory-agent performance over outcome-only GRPO 
and a comparable memory-agent RL baseline, ReMemR1.

Further analyses reveal three key properties of effective final-memory rewards: they should operate as 
positive memory shaping over successful trajectories, be normalized before reward composition, 
and be conditioned on the ground-truth answer rather than on the query alone. 
Overall, our results suggest that answer-conditioned information gain provides a more direct 
and effective supervision signal for learning useful final memories in chunk-wise long-context RL.

\section*{Limitations}

Despite these promising results, our current study still has several limitations.

First, our experiments use a limited training subset and a relatively small base model. 
This design reflects both the high computational cost of long-context reinforcement learning 
and our focus on controlled evaluation of the proposed reward design rather than large-scale 
benchmark optimization. 
Scaling InfoMem to larger models and substantially larger training corpora remains an 
important direction for future study.

Second, this work focuses specifically on chunk-wise long-context memory agents. 
The proposed reward is designed for settings in which the model sequentially processes 
context chunks, maintains an explicit memory state, and generates the final answer from 
the resulting final memory. 
Its applicability to other long-context paradigms, such as retrieval-only systems or 
full-context single-pass models, remains unexplored.

Third, the current reward is defined only at the final step. 
Although GRPO propagates the resulting trajectory-level advantage to all generated tokens, 
the reward itself evaluates only the final memory and final answer. 
Extending answer-conditioned information gain toward intermediate memory states and 
step-wise process supervision remains future work.

Potential risks should also be considered. Since InfoMem encourages final memories that 
increase support for a target answer, erroneous answers or biased source documents may 
lead the model to preserve and amplify misleading evidence. The answer-conditioned reward 
may also be over-optimized toward memories that are highly associated with the expected 
answer while omitting important qualifications from the original context. These risks are 
especially relevant in high-stakes long-document applications, such as legal, medical, or 
financial analysis, where compressed memory states should not replace source-document 
verification or human review.

\bibliography{custom}

\appendix

\input{appendix}
\end{document}

%% file: appendix.tex
\section{Implementation Details}
\label{app:implementation-details}

\subsection{Model, Training Data, and Compute}
\label{app:model-training-compute}

All training runs use Qwen2.5-1.5B-Instruct~\citep{qwen2.5technicalreport} 
as the base model and the corresponding 
Qwen2.5 tokenizer. All compared methods are initialized from the same base checkpoint 
and are trained under the same chunk-wise memory-agent framework.
We implement RL fine-tuning with the veRL framework~\citep{sheng2025hybridflow}. 

We train on RULER-HotpotQA~\citep{yu2025memagentreshapinglongcontextllm}, 
a long-document QA dataset constructed by applying the 
RULER~\citep{hsiehruler} long-context generation paradigm to HotpotQA~\citep{yang-etal-2018-hotpotqa}. 
Each example contains a query, 
a ground-truth answer, and a long context consisting of 200 documents. The original 
training split contains 32,768 examples, totaling approximately 973M tokens, and 
the validation split contains 128 examples, totaling approximately 4.1M tokens. 
To focus on controlled reward-design evaluation rather than large-scale benchmark 
optimization, we downsample the training split to 512 examples, corresponding to 
approximately 15.2M tokens, while keeping the validation split unchanged.

Table~\ref{tab:appendix-training-config} summarizes the main training configuration. 
All main comparison methods use the same data, base model, rollout number, optimization 
budget, decoding configuration, and compute budget. Each training run requires 
approximately 440 GPU-hours on 16 NVIDIA H20 GPUs.

\begin{table}[t]
\centering
{\small
\setlength{\tabcolsep}{4pt}
\renewcommand{\arraystretch}{1.1}
\begin{tabular}{@{}p{0.48\columnwidth}p{0.42\columnwidth}@{}}
\hline
\textbf{Hyperparameter} & \textbf{Value} \\
\hline
Base model & Qwen2.5-1.5B-Instruct \\
Tokenizer & Qwen2.5 tokenizer \\
Training dataset & RULER-HotpotQA \\
Training examples & 512 \\
Validation examples & 128 \\
Training tokens & about 15.2M \\
Validation tokens & about 4.1M \\
Documents per example & 200 \\
Algorithm & GRPO \\
Rollouts per prompt & 8 \\
Rollout temperature & 1 \\
Rollout top-$p$ & 1 \\
Training batch size & 256 \\
PPO mini-batch size & 64 \\
PPO epochs & 1 \\
Clip epsilon & 0.2 \\
Learning rate & $1\times10^{-6}$ \\
Warmup steps & 2 \\
KL loss weight & 0.001 \\
Training steps & 120 \\
Chunk size & 5000 tokens \\
Maximum memory length & 1024 tokens \\
InfoMem coefficient $\beta$ & 0.2 \\
GPU-hours per run & about 440 \\
\hline
\end{tabular}
}
\caption{Training configuration used for the main comparison experiments. Decoding, 
maximum input/output length, and KL-related settings are kept fixed across compared methods.}
\label{tab:appendix-training-config}
\end{table}

ReMemR1 additionally uses method-specific advantage-composition parameters. 
Since ReMemR1 retrieves previous memories and inserts them back into the recurrent 
prompt, we set its maximum prompt length to 2048 tokens. These ReMemR1-specific 
settings are kept consistent with the original ReMemR1 setting and are summarized 
in Table~\ref{tab:appendix-rememr1-config}.

\begin{table}[t]
\centering
{\small
\setlength{\tabcolsep}{6pt}
\renewcommand{\arraystretch}{1.1}
\begin{tabular}{@{}ll@{}}
\hline
\textbf{Parameter} & \textbf{Value} \\
\hline
Advantage mixing coefficient $\alpha$ & 0.8 \\
Action reweighting & false \\
Maximum prompt length & 2048 tokens \\
\hline
\end{tabular}
}
\caption{ReMemR1-specific reward and advantage parameters.}
\label{tab:appendix-rememr1-config}
\end{table}

\subsection{GRPO and Reward Implementation}
\label{app:grpo-reward-implementation}

Algorithm~\ref{alg:infomem} summarizes the reward-construction procedure used to 
interface InfoMem with GRPO. The procedure only changes the scalar reward assigned 
to each rollout; the GRPO optimizer itself is unchanged.

\begin{algorithm}[t]
\small
\caption{InfoMem Reward Construction}
\label{alg:infomem}
\begin{algorithmic}[1]
\Require Query $x$, long context $D$, ground-truth answer $y^*$, rollout group size $n$
\Ensure Rollout rewards $\{R_i\}_{i=1}^{n}$ for GRPO

\State $\{(M_i,\hat{y}_i)\}_{i=1}^{n} \gets \textsc{Rollout}(\pi_\theta, x, D, n)$

\For{$i=1,\ldots,n$}
    \State $R_{\mathrm{outcome},i} \gets 
    \mathbbm{1}\!\left[\mathrm{Norm}(\hat{y}_i)=\mathrm{Norm}(y^*)\right]$
    \State $R_i \gets R_{\mathrm{outcome},i}$
\EndFor

\State $\mathcal{S} \gets \{i : R_{\mathrm{outcome},i}=1\}$

\If{$|\mathcal{S}|>0$}
    \For{$i \in \mathcal{S}$}
        \State $r_i \gets r_{\mathrm{gain}}(x,M_i,y^*)$ \Comment{teacher-forced scoring}
    \EndFor

    \If{$|\mathcal{S}|=1$}
        \State $\tilde{r}_i \gets r_i$ for the only $i\in\mathcal{S}$
    \Else
        \State $\mu_{\mathcal{S}} \gets \mathrm{mean}(\{r_i:i\in\mathcal{S}\})$
        \State $\sigma_{\mathcal{S}} \gets \mathrm{std}(\{r_i:i\in\mathcal{S}\})$
        \For{$i \in \mathcal{S}$}
            \State $\tilde{r}_i \gets (r_i-\mu_{\mathcal{S}})/(\sigma_{\mathcal{S}}+\epsilon)$
        \EndFor
    \EndIf

    \For{$i \in \mathcal{S}$}
        \State $R_i \gets R_i + \beta \tilde{r}_i$
    \EndFor
\EndIf

\State \Return $\{R_i\}_{i=1}^{n}$
\end{algorithmic}
\end{algorithm}

The reward computation is detached from the policy-gradient path. After reward 
composition, the resulting trajectory-level advantage is assigned to all generated 
tokens in the rollout, including both memory-update tokens and final-answer tokens. 
The normalized matching rule used to compute $R_{\mathrm{outcome}}$ in 
Algorithm~\ref{alg:infomem} is used only for training-time rewards; benchmark evaluation 
uses the task-specific metrics in Section~\ref{sec:benchmarks}.

\subsection{Training-time Outcome Reward}

During training, the sparse outcome reward is computed by normalized boxed-answer matching 
following the default settings in veRL framework~\citep{sheng2025hybridflow}. 
For each rollout, we first keep the last 300 characters of the generated solution and convert them to lowercase. 
We then extract the answer from the last occurrence of either \texttt{\textbackslash boxed\{...\}} 
or \texttt{\textbackslash boxed } in this suffix, as the training prompts explicitly require the model to 
place the final answer inside a \texttt{\textbackslash boxed\{\}} expression.
If no boxed answer is found, the rollout is assigned zero outcome reward.

The extracted answer is compared against a list of ground-truth answers. 
A rollout is marked successful if the extracted answer matches any ground-truth answer after normalization:
\[
R_{\mathrm{outcome}} =
\mathbbm{1}\left[
\exists y \in \mathcal{Y}^*: 
\operatorname{Norm}(\hat y)=\operatorname{Norm}(y)
\right],
\]
where $\mathcal{Y}^*$ denotes the set of acceptable ground-truth answers. 
The normalization follows the string normalization used in the Hendrycks 
MATH evaluation script from EleutherAI's lm-evaluation-harness~\citep{eval-harness}. 
It removes line breaks, inverse spaces, dollar signs, percentage symbols, \texttt{\textbackslash left}/\texttt{\textbackslash right}, degree markers, and whitespace; normalizes \texttt{tfrac}/\texttt{dfrac} to \texttt{frac}; normalizes escaped backslashes; and canonicalizes decimal forms such as \texttt{.5} to \texttt{0.5}. 
The normalized strings are then compared by exact matching.

\section{Synthetic Hallucinated-Evidence Diagnostic}
\label{app:synthetic-hallucination}

\subsection{Dataset Construction}
\label{app:synthetic-dataset-construction}

We construct the hallucinated-evidence diagnostic from SQuAD~\citep{rajpurkar-etal-2016-squad}. 
Each original example contains a question, a supporting context, and a ground-truth 
answer. To reduce the influence of parametric knowledge, we retain only examples for 
which the scoring model fails to answer correctly without context but succeeds when 
conditioned on the original supporting context. This filtering makes the diagnostic 
focus on context-dependent evidence use rather than memorized answer recall.

For each retained QA pair, we build a three-context candidate set consisting of the 
original supporting context and two synthetic hallucinated contexts. The hallucinated 
contexts are generated with Gemini 3 Flash preview~\citep{google2026gemini3flashpreview}. 
They are required to preserve the topic, discourse structure, and surface style of the 
original passage while replacing answer-critical facts so that the resulting text no 
longer supports the ground-truth answer. This creates a controlled setting in which 
lexical and topical similarity alone is insufficient for identifying the truly 
answer-supporting context.
After construction, each diagnostic instance contains one positive 
context and two hallucinated negative contexts, and each evidence score is evaluated by 
ranking these three candidates for the same question-answer pair.

\subsection{Hallucinated Context Generation Prompt}
\label{app:hallucination-generation-prompt}

We use Gemini 3 Flash preview to generate two hallucinated contexts for each retained 
QA pair. The prompt provides the question, the ground-truth answer, and the original 
supporting context. The generation is constrained to keep the passage topically and 
stylistically similar to the original context while corrupting answer-critical facts, 
so that the generated contexts remain plausible distractors but no longer support the 
ground-truth answer. The prompt template is shown below.

\begingroup
\small
\begin{verbatim}
You are given a QA sample.
Generate two hallucinated contexts.

Requirements:
1. Keep each context relevant to the question.
2. Preserve the topic, style, and structure of
   the reference context as much as possible.
3. Modify key supporting facts so that the
   context does not support the gold answer.
4. Do not change the question.
5. Do not remove all relevant information.
6. Do not reveal the gold answer directly.
7. Do not add meta text, warnings, or labels
   indicating that the context is false.

Question:
{question}

Gold answer:
{gold_answer}

Reference context:
{context}

Return format:
{
  "hallucinated_context_1": "...",
  "hallucinated_context_2": "..."
}
\end{verbatim}
\endgroup

We conduct manual spot checks to verify the quality of the generated hallucinated 
contexts, ensuring that they remain fluent and plausible while no longer supporting the 
ground-truth answer.

\subsection{Evidence Score Computation}
\label{app:evidence-score-computation}

\subsubsection{Information-gain Score}

For each candidate context $c$, we compute the answer-conditioned information-gain 
score with the same Qwen2.5-1.5B-Instruct model used in training:
\begin{equation}
\begin{aligned}
  r_{\mathrm{gain}}(x,c,y^*)
  &=
  \frac{1}{|y^*|}\log P_\theta(y^* \mid x,c)
  \\
  &\quad -
  \frac{1}{|y^*|}\log P_\theta(y^* \mid x,\emptyset).
\end{aligned}
\end{equation}
The two likelihood terms are computed under teacher forcing over the ground-truth 
answer tokens. The null-context term uses the same query $x$ but replaces the candidate 
context with an empty context. No answer sampling or decoding is used for this score.

\subsubsection{Embedding Similarity}

For embedding-based scores, we compare each candidate context against a 
question-answer query representation. Specifically, we encode the candidate context $c$ 
and the template \textit{``Question: \textnormal{\texttt{\{QUESTION\}}} Answer: 
\textnormal{\texttt{\{ANSWER\}}}''} with the same embedding 
model, then compute cosine similarity between the two embeddings. We evaluate this 
procedure with three open-source embedding models: Qwen3-Embedding-0.6B~\citep{zhang2025qwen3embeddingadvancingtext}, 
BGE-M3~\citep{chen-etal-2024-m3}, and 
multilingual-E5-large-instruct~\citep{wang2024multilinguale5}. Scores are computed independently for each embedding 
model.

\subsubsection{Attention-based Scores}

For attention-based scores, we use the same Qwen2.5-1.5B-Instruct model and feed the 
candidate context, question, and ground-truth answer into the model. We extract 
attention from the final attention layer. Since the layer contains multiple attention 
heads, we first average attention weights across heads. Let $A_{u,v}$ denote the resulting 
head-averaged final-layer attention from token $u$ to token $v$, $\mathcal{Y}$ denote 
the answer-token positions, and $\mathcal{C}$ denote the candidate-context token positions.

\textsc{Attn-Mass} measures the average total attention assigned from answer tokens to 
candidate-context tokens:
\begin{equation}
  s_{\mathrm{mass}}
  =
  \frac{1}{|\mathcal{Y}|}
  \sum_{u\in\mathcal{Y}}
  \sum_{v\in\mathcal{C}} A_{u,v}.
\end{equation}
\textsc{Attn-Top1} measures the average strongest context-token attention for each 
answer token:
\begin{equation}
  s_{\mathrm{top1}}
  =
  \frac{1}{|\mathcal{Y}|}
  \sum_{u\in\mathcal{Y}}
  \max_{v\in\mathcal{C}} A_{u,v}.
\end{equation}
Both attention scores are higher-is-better and are computed without using the generated 
answer.

\subsection{Metrics: MRR and Z-score SNR}
\label{app:synthetic-metrics}

For each question, the candidate set contains one true supporting context and two 
hallucinated contexts. Since all scores are higher-is-better, we sort the three 
candidates in descending score order and record the rank of the true supporting context. 
Given $N$ diagnostic examples, mean reciprocal rank is computed as
\begin{equation}
  \mathrm{MRR}
  =
  \frac{1}{N}
  \sum_{q=1}^{N}
  \frac{1}{\operatorname{rank}_{\text{supporting context}}}.
\end{equation}
MRR measures whether a score ranks the true supporting context ahead of hallucinated 
contexts.

We also compute a Z-score signal-to-noise ratio (SNR) to measure the stability of the 
true-versus-hallucinated separation. For each question $q$, let 
$s_q^+$ denote the score of the true context and $s_{q,1}^-,s_{q,2}^-$ denote the two 
hallucinated-context scores. We first standardize the three scores within the same 
question:
\begin{equation}
  z_{q,j}
  =
  \frac{s_{q,j}-\mu_q}{\sigma_q+\epsilon},
  \quad j\in\{+,1,2\},
\end{equation}
where $\mu_q$ and $\sigma_q$ are computed over the three candidate scores for question 
$q$, and $\epsilon$ is used for numerical stability. We then compute two standardized 
margins for each question:
\begin{equation}
  \Delta_{q,k}
  =
  z_q^+ - z_{q,k}^-,
  \quad k\in\{1,2\}.
\end{equation}
Pooling all $2N$ margins into $\Delta$, the reported SNR is
\begin{equation}
  \mathrm{SNR}
  =
  \frac{\operatorname{mean}(\Delta)}
  {\operatorname{std}(\Delta)}.
\end{equation}
A higher SNR indicates that the true context is separated from hallucinated contexts 
with a larger and more stable standardized margin, making the score more suitable as a 
reward signal.

\section{Evaluation Protocols}
\label{app:evaluation-protocols}

\subsection{Benchmark Summary}
\label{app:benchmark-summary}

Table~\ref{tab:appendix-benchmark-summary} summarizes the evaluation benchmarks used in 
Section~\ref{sec:evaluation}. InfoMem and the outcome-only GRPO baseline are evaluated 
under the same chunk-wise memory-agent framework, with the same memory budget, decoding 
configuration, benchmark subset, and metric for each benchmark. ReMemR1 is evaluated 
under its original callback-retrieval framework, described in 
Appendix~\ref{app:rememr1-framework}, to preserve train-test consistency.

\begin{table*}[t]
\centering
{\small
\setlength{\tabcolsep}{3pt}
\renewcommand{\arraystretch}{1.15}
\begin{tabular}{@{}p{0.14\textwidth}p{0.16\textwidth}p{0.09\textwidth}p{0.16\textwidth}p{0.12\textwidth}p{0.09\textwidth}p{0.14\textwidth}@{}}
\hline
\textbf{Benchmark} & \textbf{Task Type} & \textbf{Samples} & \textbf{Sample Context} & \textbf{Metric} & \textbf{Judge} & \textbf{Subset} \\
\hline
MRCR-8needle & NIAH,8 needles & 100 & 4K$\sim$1M tokens & Seq. match & N/A & 100 / 800 \\
RULER synthetic QA & Sparse QA & 128 & 262K tokens & F1 score & N/A & 262K subsets \\
CorpusQA & Corpus QA & 329 & 128K tokens & LLM judge & Kimi-K2.6 & 128K subset \\
LongMemEval & Dialogue QA & 500 & 115K tokens & LLM judge & Kimi-K2.6 & LongMemEval-S \\
\hline
\end{tabular}
}
\caption{Summary of evaluation benchmarks and metrics.}
\label{tab:appendix-benchmark-summary}
\end{table*}

MRCR-8needle~\citep{openai2025mrcr} is particularly challenging in the chunk-wise memory-agent setting because 
the model cannot directly attend over the full context. Instead, it must preserve multiple 
sparse targets through compressed sequential memory updates. This setting is difficult 
for a 1.5B model under an 8-needle retrieval task, so absolute sequence-match scores are 
low across methods. The comparison remains controlled because all methods use the same 
model scale, memory length, downsampled subset, and evaluation protocol.

For RULER synthetic QA~\citep{yu2025memagentreshapinglongcontextllm}, 
we use the 262,144-token evaluation setting and report F1. 
CorpusQA~\citep{lu2026corpusqa10milliontoken} is evaluated on the 128K-token subset, 
and LongMemEval~\citep{wulongmemeval} is evaluated on the 
115K-token LongMemEval-S subset. For CorpusQA and LongMemEval, we use LLM-as-judge 
evaluation with Kimi-K2.6; the judging protocol is described in 
Appendix~\ref{app:llm-as-judge-protocol}. For MRCR-8needle, we use the same fixed 
100-example subset for all methods to reduce evaluation cost. All model generations 
during benchmark evaluation are served with vLLM~\citep{DBLP:conf/sosp/KwonLZ0ZY0ZS23}.
Each evaluation is run once. Following the original ReMemR1 paper and code, ReMemR1 
is evaluated with temperature sampling at $t=0.7$, whereas all other evaluations use 
greedy sampling.
The evaluation datasets are released under permissive MIT or Apache-2.0 licenses and 
can be used for academic research.

\subsection{ReMemR1 Callback-Retrieval Framework}
\label{app:rememr1-framework}

ReMemR1 augments the standard chunk-wise memory-agent framework with callback retrieval 
over previously generated memories. At each chunk, the model maintains a current memory 
and a callback query. The callback query is used to retrieve a relevant historical memory 
from the memories generated in earlier steps, and the retrieved memory is inserted into 
the next recurrent prompt together with the current memory, the current chunk, and the 
original question. The model then generates an updated memory and a new callback query 
for the next step. At the final step, ReMemR1 similarly uses the accumulated memory state 
and callback-retrieved historical memory to produce the answer.

Because this callback mechanism changes the inference framework rather than only the 
reward function, we do not force ReMemR1 into the pure chunk-wise setting used by InfoMem. 
Instead, we keep the shared settings aligned where applicable, including the base model, 
training data, chunk size, rollout decoding configuration, and benchmark evaluation 
protocol, while preserving ReMemR1-specific callback and advantage settings.

\subsection{LLM-as-Judge Protocol}
\label{app:llm-as-judge-protocol}

CorpusQA and LongMemEval are evaluated with the same binary answer-equivalence 
judging protocol. For each example, the judge receives the query, the ground-truth 
answer candidates, and the model answer. It does not receive the method name or any 
training configuration. The judge returns a binary label, where 1 denotes a correct 
answer and 0 denotes an incorrect answer. The reported score is the percentage of 
examples labeled correct.

We use Kimi-K2.6 as the judge model and apply the following prompt:

{\small
\begin{verbatim}
SYSTEM_PROMPT = """You are a strict
answer-equivalence judge.
Return 1 only when the model answer fully
contains one ground-truth candidate with
exactly the same meaning: nothing meaningful
is missing and nothing meaningful is added.
Ignore only semantically empty surface
differences, such as articles like "the",
punctuation, spaces, line breaks, case, LaTeX,
or boxing / markup symbols.
If the answer is missing any part of the
ground truth, adds any meaningful content,
or you are unsure, return 0.

First write a brief reason. Then write the
final binary judgment inside
<answer></answer>.
The <answer> tag must contain exactly one
character: 0 or 1."""

USER_TEMPLATE = """Query:
{query}

Ground truth candidates:
{ground_truth_candidates}

Model answer:
{answer}

Does the model answer fully contain any
ground-truth candidate with exactly the same
meaning, allowing only semantically empty
surface characters such as "the", punctuation,
spaces, case differences, or boxing / markup
symbols?

Output format:
Reason: <brief reason>
<answer>0 or 1</answer>"""
\end{verbatim}
}

The final decision is parsed from the binary value inside the 
\texttt{<answer>} tag. This protocol treats paraphrases and minor formatting 
differences as correct only when the predicted answer preserves the complete 
meaning of one ground-truth candidate. Answers are marked incorrect if they omit 
essential information, contradict the ground truth, add meaningful unsupported 
content, or cannot be confidently judged as equivalent.

\subsection{Human Calibration of LLM-as-Judge}
\label{app:human-calibration}

We further conduct human calibration to assess the reliability of the LLM judge. 
We hired an undergraduate annotator and presented each example with the query, the 
ground-truth answer candidates, and the model answer. The annotator assigned the same 
binary correctness label as the LLM-as-judge protocol, and we compared the human labels 
with the LLM-judge outputs. We explicitly informed the annotator that the annotations 
would be used for anonymized academic research. The annotator was paid at an hourly rate 
ten times the local minimum wage.

For calibration, we inspect three CorpusQA result files, corresponding to the initial 
instruct model, the Outcome-only GRPO baseline, and InfoMem, as well as one LongMemEval 
result file from the Outcome-only GRPO baseline. This verification shows high agreement 
with human judgments, with 99.5\% agreement on CorpusQA and 96.8\% agreement on 
LongMemEval.

\subsection{Checkpoint Selection Protocol}
\label{app:checkpoint-selection}

For the main comparison experiments, all RL methods are trained for 120 optimization 
steps under the same training budget. During training, we evaluate checkpoints every 
2 optimization steps on the validation split described in Appendix~\ref{app:model-training-compute}. 
The final checkpoint reported for each method is selected according to the highest 
validation accuracy under this fixed validation protocol. If multiple checkpoints obtain 
the same validation accuracy, we select the earlier checkpoint to favor better 
generalization.

This selection rule is fixed before benchmark comparison and is applied consistently 
to all training methods. It prevents selecting checkpoints based on test-set 
performance, while still accounting for the instability of long-context RL fine-tuning. 
Diagnostic variants are evaluated with the same validation-based selection principle 
when benchmark scores are reported.

\section{Prompt Templates}
\label{app:prompt-templates}

The memory-update and final-answer templates follow the MemAgent setting~\citep{yu2025memagentreshapinglongcontextllm}. 
We use the following template for chunk-wise memory updates:

{\small
\begin{verbatim}
TEMPLATE = """
You are presented with a problem, a section of an 
article that may contain the answer to the problem, 
and a previous memory. Please read the provided 
section carefully and update the memory with the 
new information that helps to answer the problem. 
Be sure to retain all relevant details from the 
previous memory while adding any new, useful 
information.

<problem> 
{prompt}
</problem>

<memory>
{memory}
</memory>

<section>
{chunk}
</section>

Updated memory:
"""
\end{verbatim}
}

The final answer is generated with the following template:

{\small
\begin{verbatim}
TEMPLATE_FINAL_BOXED = """
You are presented with a problem and a previous 
memory. Please answer the problem based on the 
previous memory and put the answer in \\boxed{{}}.

<problem> 
{prompt}
</problem>

<memory>
{memory}
</memory>

Your answer:
"""
\end{verbatim}
}

The information-gain reward $r_{\mathrm{gain}}$ uses the same final-answer template: 
the ground-truth answer is appended after \texttt{Your answer:} and scored by 
teacher-forced per-token log-likelihood, with either the final memory or an empty 
memory field. 

QueryPMI instead uses the following query-conditioned template:

{\small
\begin{verbatim}
TEMPLATE_QUERY_PMI = """
Based on the previous memory,

<memory>
{memory}
</memory>

we can answer the Query: """
\end{verbatim}
}

\section{Additional Training Curves and Diagnostics}
\label{app:training-curves-diagnostics}

\subsection{Main Training Curves}
\label{app:main-training-curves}

\begin{figure}[!htbp]
  \centering
  \includegraphics[width=0.88\columnwidth]{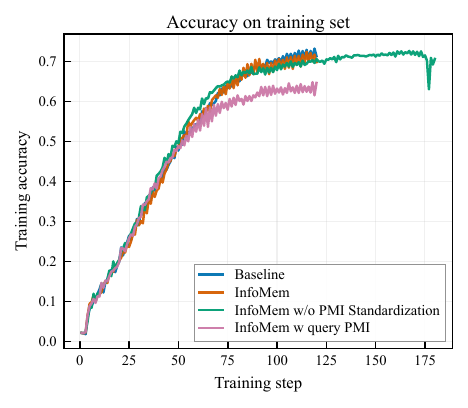}
  \caption{Training curves for the main comparison runs and diagnostic variants.}
  \label{fig:appendix-training-curve}
\end{figure}

\begin{figure}[!htbp]
  \centering
  \includegraphics[width=0.88\columnwidth]{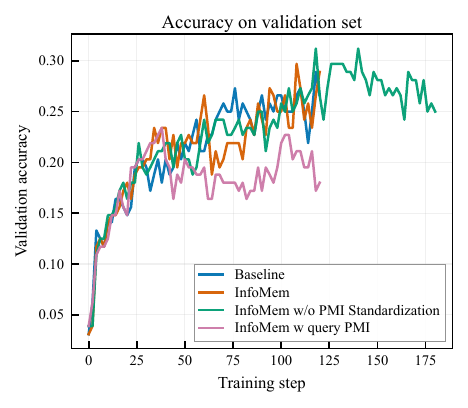}
  \caption{Validation curves for the main comparison runs and diagnostic variants.}
  \label{fig:appendix-validation-curve}
\end{figure}

\begin{figure*}[t]
  \centering
  \begin{minipage}{0.48\textwidth}
    \centering
    \includegraphics[width=\linewidth]{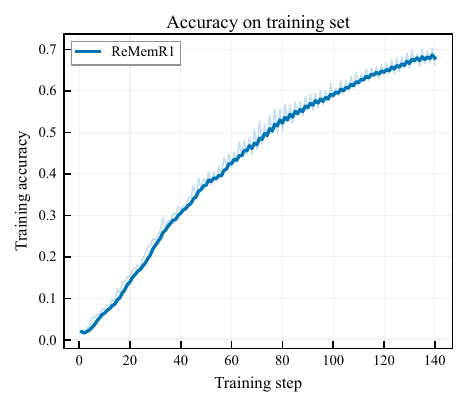}
    \textbf{(a)} Training accuracy.
  \end{minipage}
  \hfill
  \begin{minipage}{0.48\textwidth}
    \centering
    \includegraphics[width=\linewidth]{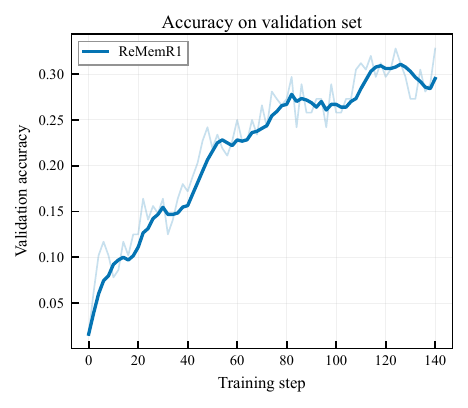}
    \textbf{(b)} Validation accuracy.
  \end{minipage}
  \caption{ReMemR1 training dynamics under its callback-retrieval chunk-wise framework. 
  Both curves are smoothed with a sliding window of 5.}
  \label{fig:appendix-rememr1-curves}
\end{figure*}

Figures~\ref{fig:appendix-training-curve} and~\ref{fig:appendix-validation-curve} 
report the training and validation curves for the main comparison runs and diagnostic 
variants. All main comparison methods are trained for 120 optimization steps under 
the same data, model, rollout number, and validation protocol. Checkpoints are 
evaluated every 2 optimization steps, and the reported benchmark checkpoint is 
selected according to the validation protocol described in 
Appendix~\ref{app:checkpoint-selection}.

The no-standardization ablation was additionally continued 
beyond 120 steps, because its curve was still increasing near the end of the original 
training budget. After continuing the run, the best checkpoint was still step 118, 
which lies within the original 120-step budget. Therefore, the reported benchmark 
result for this ablation remains comparable to the other methods under the fixed 
checkpoint-selection protocol.

ReMemR1 uses a callback-retrieval chunk-wise framework rather than the pure chunk-wise 
framework used by InfoMem and the outcome-only GRPO baseline. We therefore report its 
training dynamics separately in Figure~\ref{fig:appendix-rememr1-curves}, instead of 
overlaying them with the main comparison curves. Because of this framework difference, 
ReMemR1 is also trained for more than 120 optimization steps.